# Any equation is a forest: Symbolic genetic algorithm for discovering open-form partial differential equations (SGA-PDE)

Yuntian Chen[1], Yingtao Luo[2], Qiang Liu[3,4], Hao Xu[5], and Dongxiao Zhang[6,*]

[1]Intelligent Energy Laboratory, Frontier Research Center, Peng Cheng Laboratory, Shenzhen, P. R. China
[2]Department of Computer Science, University of Washington, Seattle, WA, U.S.A.
[3]Center for Research on Intelligent Perception and Computing, Institute of Automation, Chinese Academy of Sciences, Beijing, P. R. China
[4]School of Artificial Intelligence, University of Chinese Academy of Sciences, Beijing, P. R. China
[5]BIC-ESAT, ERE, and SKLTCS, College of Engineering, Peking University, Beijing, P. R. China
[6]School of Environmental Science and Engineering, Southern University of Science and Technology, Shenzhen, P. R. China

**Abstract**
Partial differential equations (PDEs) are concise and understandable representations of domain knowledge, which are essential for deepening our understanding of physical processes and predicting future responses. However, the PDEs of many real-world problems are uncertain, which calls for PDE discovery. We propose the symbolic genetic algorithm (SGA-PDE) to discover open-form PDEs directly from data without prior knowledge about the equation structure. SGA-PDE focuses on the representation and optimization of PDE. Firstly, SGA-PDE uses symbolic mathematics to realize the flexible representation of any given PDE, transforms a PDE into a forest, and converts each function term into a binary tree. Secondly, SGA-PDE adopts a specially designed genetic algorithm to efficiently optimize the binary trees by iteratively updating the tree topology and node attributes. The SGA-PDE is gradient-free, which is a desirable characteristic in PDE discovery since it is difficult to obtain the gradient between the PDE loss and the PDE structure. In the experiment, SGA-PDE not only successfully discovered nonlinear Burgers' equation, Korteweg-de Vries (KdV) equation, and Chafee-Infante equation, but also handled PDEs with fractional structure and compound functions that cannot be solved by conventional PDE discovery methods.

## 1. Introduction

Over the past centuries, as people's understanding of the world has gradually deepened, a large amount of domain knowledge has been accumulated. Much of the domain knowledge comes from the experience of experts and the theoretical research of scholars, and is expressed as various theoretical models or empirical models mostly represented by partial differential equations (PDEs). However, many systems in practical engineering applications are too complex and irregular, resulting in complicated forms of PDEs (governing equations) describing the mapping between variables, which are difficult to derive directly from theory. Therefore, researchers often collect data through physical experiments and obtain governing equations by analyzing the experimental data. Data may also be obtained via high-resolution simulations[1-3]. The results of numerical simulations have comprehensive information to describe the problem, but due to the complexity of real-world problems, numerical simulations not only have high computational cost, but the results are also only applicable to specific scenarios at a certain scale. Therefore, it is difficult to abstract general knowledge and principles from simulation results, which limits the transfer and application of simulation results between different scenarios in practice. In real-world applications, a common aspiration is to automatically mine the most valuable and important internal principles (i.e., governing equations) from high-dimensional nonlinear (experimental or simulation) data directly through a machine learning method, so as to realize automatic knowledge discovery.

In recent years, many studies have been carried out in the field of knowledge discovery to mine governing equations from data. Although numerous important achievements have been made, knowledge discovery remains an open problem. Indeed, there are still many deficiencies in mining PDE from data, which calls for further research. At present, however, there are mainly two kinds of methods of automatic mining PDE: sparse regression and genetic algorithm.

Sparse regression is the most widely used algorithm in PDE discovery. Rudy et al. proposed PDE-FIND in 2017 and mined the Navier Stokes equation from data based on sequential threshold ridge regression (STRidge)[4]. Chang et al. used LASSO (least absolute shrinkage and selection operator) to mine underground seepage governing equation from data in 2019[5]. In addition, Xu et al. developed DL-PDE based on ridge regression to discover PDE from discrete and noisy data in 2019[6]. LASSO and STRidge are the main sparse regression algorithms in solving the problem of PDE discovery with constant coefficients based on the given candidate set[4, 7, 8], but the space that these methods can explore is limited to the space covered by the candidate set. In other words, these methods require the user to determine the approximate form of the governing equation in advance, and then give all possible differential operators as the function terms in the candidate set. It is impossible to find the function terms that do not exist in the candidate set from the data in these methods. In addition to the sparse-regression-based methods, Long et al. proposed PDE-net to accurately predict dynamics of complex systems and to elucidate the underlying hidden PDE models in 2018[9]. PDE-net is essentially equivalent to a multivariate regression, in which the differential operators are represented by convolution kernels in the learned network, and they are also determined in advance. However, the governing equations to be mined often have complex forms in the scenario in which the PDE discovery method is needed, and it is challenging to generate a complete candidate set containing all of the possible function terms in practice. Therefore, although sparse regression achieves good performance in some simple problems, it cannot find the solution not included in the candidate set, which constrains the application of these sparse-regression-based methods in practice.

In order to solve the problem that all possible candidates (function terms) need to be given in advance in sparse regression, Maslyaev et al. proposed EPDE based on the evolutionary method in 2019[10], and Xu et al. developed DLGA-PDE based on the genetic algorithm in 2020[11], which uses crossover and mutation to expand the candidate set, so that the PDE can be automatically found without determining all of the candidates in advance. EPDE discovers Burgers' equation, wave equation, and Korteweg-de Vries (KdV) equation from observed data even if the candidate set is incomplete. Moreover, DLGA-PDE successfully handles the problem of Burgers' equation, Chafee-Infante equation, and KdV equation. However, because the basic units of variation (gene fragments) in EPDE and DLGA-PDE are the derivative of different orders, it can only produce the terms generated by direct interactions between the basic derivatives, which indicates that the exploration space of EPDE and DLGA-PDE is limited to the interaction space of the given candidate set. In other words, the current evolutionary-strategy-based methods are still unable to mine open-form PDEs from data (e.g., the PDE with compound function or fractional structure). The limitation of sparse regression and genetic algorithm lies in that the representation of PDE is based on the given candidates (function terms) or gene fragments (basic derivatives), which can only carry out simple interactive operations, such as addition, subtraction, and multiplication. These basic representation units with limited expression ability constrain the potential equation structure that the algorithm can find, and thus it is challenging to discover the open-form PDEs from observations. In addition to sparse regression and genetic algorithms, researchers have also attempted other solutions. For instance, Long et al. proposed PDE-Net 2.0[12] to discover PDEs from observation without the assumption about the general type of the PDE by introducing a symbolic neural network based on the idea of PDE-net[9]. PDE-Net 2.0 has the potential to uncover the hidden governing equation of the observed dynamics. However, the symbolic neural network in PDE-net 2.0 only has addition and multiplication, which is too restrictive to find PDEs with complex structures (i.e., open-form PDE discovery). The different models are compared in Table 1.

**Table 1.** Comparison of different PDE discovery models.

| Model | Method | Candidate set | Representation/optimization units |
|---|---|---|---|
| PDE-FIND[4] | Sparse regression | Close | Function terms |
| DL-PDE[6] | Sparse regression | Close | Function terms |
| EPDE[10] | Genetic algorithm | Limited | Derivatives |
| DLGA-PDE[11] | Genetic algorithm | Limited | Derivatives |
| SGA-PDE | Symbolic mathematics and genetic algorithm | Open | Operands and operators |

Considering the aforementioned problems, a salient question is: is there any flexible representation that is able to handle any open-form PDE, and is it possible to build an effective optimization algorithm for this representation to discover the PDE that fits the observations? By answering the above question, the analytical form behind the physical process can be recovered from the data, and the governing equation can be mined via machine learning methods.

From the perspective of symbolic mathematics, any equation can be expressed as a binary tree[13], which might be a feasible representation of open-form PDE. In this study, we propose the symbolic genetic algorithm (SGA-PDE), in which the symbolic binary trees are taken as a representation of function terms and each open-form PDE is regarded as a forest, and then we build a genetic algorithm specially designed for trees to mine the most suitable PDE directly from experimental data.

This paper comprises four sections. In section 2, we introduce SGA-PDE in detail. Section 2.1 introduces the method of flexible transformation between any open-form PDE and forest with binary trees. In section 2.2, the tree-based random generation method of open-form PDEs is discussed. Section 2.3 proposes a specially designed genetic algorithm for tree structure. In section 3, the performance of SGA-PDE is verified by experiments, and SGA-PDE successfully discovers five different governing equations from observations, including Burgers' equation, KdV equation,

Chafee-Infante equation, and two complex PDEs with compound function and fractional structure. Finally, this study is summarized and discussed in section 4.

## 2. Methodology

In this work, we aim to discover the open-form PDEs with the following form:

$$u_t = \Phi(u,x) \cdot \xi \tag{1}$$

with

$$\Phi(u,x) = [f_1(u,x), f_2(u,x), f_3(u,x), \ldots\ldots, f_i(u,x), f_{i+1}(u,x), \ldots\ldots] \tag{2}$$

where $u$ denotes the solution or observation of the considered problem; $x$ denotes the independent variable in the PDE; $f_i(u,x)$ represents a function term related to $u$ and $x$; and $\xi$ denotes the coefficient vector, which is always sparse and has many zero elements in practice.

The PDE discovery models based on sparse regression (e.g., PDE-FIND and DL-PDE) require $\Phi(u,x)$ to include all of the function terms that may appear in the PDE of the considered problem. In addition, conventional genetic algorithm models (e.g., DLGA-PDE) assume that all terms in the PDE of the considered problem can be generated by direct interactions between $f_i(u,x)$ in the $\Phi(u)$. Due to the limited function terms in $\Phi(u,x)$, the above methods cannot discover complex open-form PDEs in practice. In order to solve these problems, we need to ensure that any form of function terms can be quickly generated as $f_i(u,x)$ in $\Phi(u,x)$. Therefore, the key to directly mining governing equations from data is to solve the problem of representation and optimization (Fig. 1). Specifically, we need to express any form of equations in a computable way (i.e., generate $\Phi(u,x)$), and to evaluate their performance by measuring the fitness of the discovered equation to the data, so as to iteratively optimize the equations (i.e., determine $\xi$).

This study proposes a symbolic genetic algorithm (SGA-PDE) to achieve this goal, which first uses symbolic mathematics to express the open-form PDEs as flexible binary trees, and then uses a genetic algorithm specially designed for trees to optimize the equation. Because the search space of binary trees is infinite, however, it is difficult to optimize the equation structure through sparse regression. This type of open-domain optimization problem is similar to the problem of neural architecture search (NAS)[14]. The main methods for solving NAS problems include: genetic algorithms[15], gradient-based methods[16, 17], and reinforcement learning[18]. Nevertheless, it is difficult to construct a method for effectively calculating the gradient between the loss and the structure of the equation, which restricts the application of gradient-based methods and reinforcement learning in PDE discovery. Therefore, this study uses the gradient-free genetic algorithm to optimize the symbolic mathematical representation of PDEs. It should be mentioned that the $\Phi(u,x)$ is iteratively updated in SGA-PDE, which indicates that any form of function terms can be generated.

The main components of SGA-PDE are shown in Fig. 1. The yellow boxes on the left are the function terms in the form of binary trees, which constitute the candidate set of the open-form PDEs. The green boxes on the right represent the genetic algorithm specially designed to optimize the binary trees. The red boxes at the bottom represent the performance assessment of the discovered PDE. SGA-PDE treats each PDE as a forest (i.e., $\Phi(u,x)$), where each binary tree corresponds to a function term (i.e., $f_i(u,x)$), and the trees in the forest are connected by weighted addition (i.e., $\Phi(u,x) \cdot \xi$).

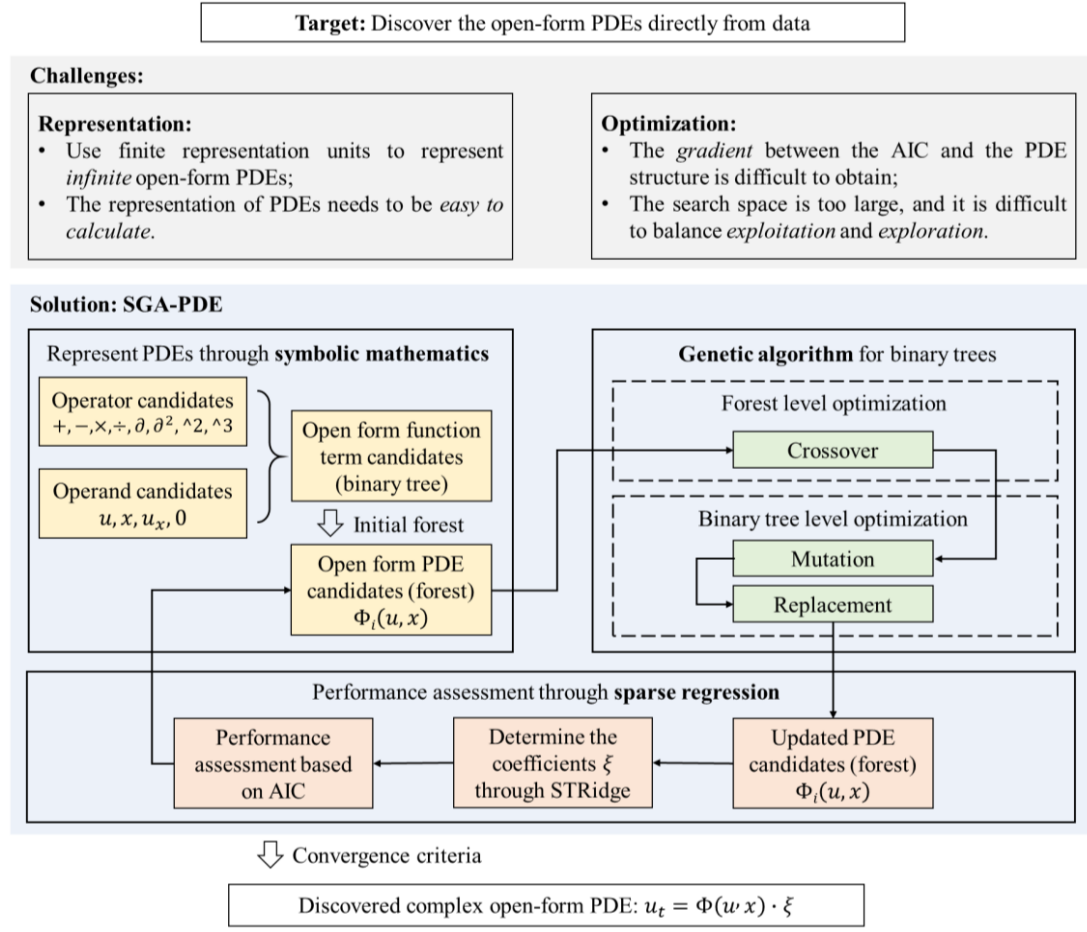

**Figure 1.** Flow chart of the symbolic genetic algorithm (SGA-PDE).

The SGA-PDE can be summarized as the following four steps: (1) generate a group of independent binary trees; (2) optimize the group of trees by changing the tree tropology and node attributes; (3) use sparse regression to determine the coefficient of each binary tree in the forest, and evaluate the fitness between the discovered PDE in the current step and the observations; and (4) iteratively optimize the topology and node attributes of the trees in the forest according to the fitness. It should be mentioned that if prior knowledge is available in the considered problem, then we can add preset candidates into the candidate set to reduce the difficulty of optimization (e.g., the diffusion terms can be preset in fluid problems according to expert experiences). However, in this study, we attempt to mine the PDEs without any priori information of the possible expression.

In the following, we will introduce the major components of SGA-PDE: the open-form PDE representation method based on symbolic mathematics (section 2.1 and section 2.2); the genetic algorithm for binary trees (section 2.3); and the performance assessment of the generated PDE (section 2.4).

**2.1 Symbolic mathematical representation of open-form PDEs**

A large number of governing equations in practice are in the form of PDEs related to time and space. This study explores automatic mining algorithms for open-form PDEs, where the key issue is how to represent the diverse open-form PDEs in a flexible manner. The conventional method to solve the PDE mining or PDE discovery problem is to determine in advance all of the candidates that may appear in the PDE, such as the derivatives of different orders, and then find the solution through sparse regression. This method can only be optimized in a closed candidate set, and cannot generate function terms outside of the fixed candidate set. Therefore, it is difficult to generate function terms with composite functions or fractional structures, as shown on the left side of Fig. 2.

In order to avoid the restriction of the closed candidate set, SGA-PDE represents open-form PDEs by refining the basic components of the equations, i.e., transforming the representation units of the equations from the function term level to the operator and operand level. Specifically, this study defines operations involving two objects as double operators (e.g., addition, subtraction, multiplication, and division). In addition, the operations involving a single object, such as exponents, logarithms, and trigonometric functions, are defined as single operators. Operands are the independent variables and dependent variables in the equation, such as $x$, $y$, $t$, and $u$. In order to encode different PDEs, this study uses binary trees to combine operators and operands, where all leaf nodes in the tree are operands, and all internal nodes correspond to the operators. The double operators are the nodes with degree two (i.e., each node has two children), and the single operators are the nodes with degree one.

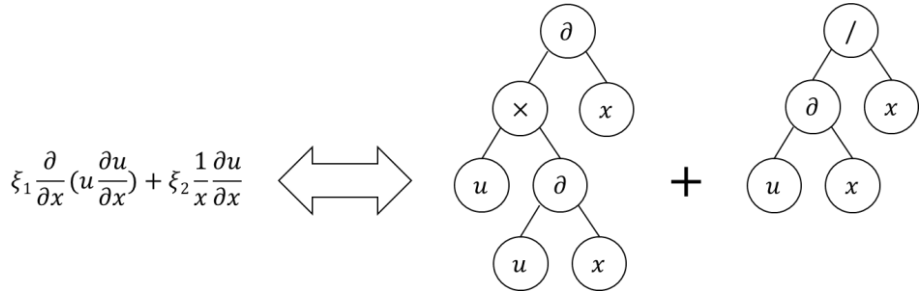

**Figure 2.** Symbolic mathematical representation of an open-form PDE.

In theory, any PDE can be transformed into a binary tree by the above method. For example, the aforementioned complex PDE with compound function and fractional structure can be written into the symbolic mathematical representation on the right side of Fig. 2. Therefore, by matching the nodes in the binary tree with the operands and operators in the PDE, any partial differential governing equation can be represented as a forest of binary trees.

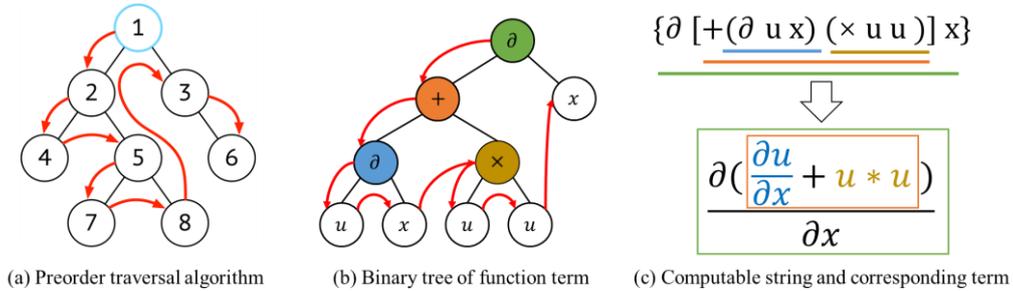

(a) Preorder traversal algorithm    (b) Binary tree of function term    (c) Computable string and corresponding term

**Figure 3.** Schematic diagram of binary tree representation and transformation method based on preorder traversal.

When performing symbolic mathematical operations, a binary tree needs to be converted into a computable string, and finally displayed as a function term that conforms to the mathematical expression habits. This study uses preorder traversal to achieve this goal (Fig. 3a). The preorder traversal starts from the root node of the binary tree. In the calculation, the left child node is processed first, and the current branch is traversed to the leaf node layer by layer. We then return to the nearest parent node whose degree is not full, enter the other branch (led by the right child node), and then traverse to the leaf node of that branch. The above process is repeated until all internal nodes of the entire binary tree are traversed. Fig. 3b shows a binary tree corresponding to the function term $\frac{\partial}{\partial x}(\frac{\partial u}{\partial x} + u \cdot u)$, and the red lines show the path of the preorder traversal. The binary tree can be converted into a computable string according to the traversal path (Fig. 3c), where each parent node and its child nodes are combined with parentheses to form a calculation unit. The parent node must be an internal node (operator) that represents a certain operation, and its child nodes (and the following branches) are the objects of the operation in the calculation unit. The different calculation units are marked with different colored underlines in Fig. 3c. Finally, in the lower part of Fig. 3c, the order of each parent node and child node in the computable string is adjusted according to mathematical expression habits. Table 2 shows three examples of explainable binary trees and forests, their corresponding computable strings, and function terms. Since the trees contained in a forest are independent of each other (i.e., there is no weighted summation between trees in the representation step), the trees are connected with "&" rather than "+".

**Table 2.** Comparison of explainable binary trees/forests, computable strings, and function terms.

| Explainable binary tree/forest | Computable strings | Function terms |
|---|---|---|
| 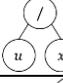 | {/ u x} | $\frac{u}{x}$ |
| 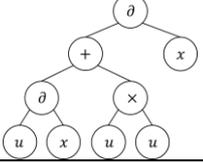 | {∂ [+ (∂ u x) (× u u)] x} | $\frac{\partial}{\partial x}\left(\frac{\partial u}{\partial x} + u \cdot u\right)$ |
| 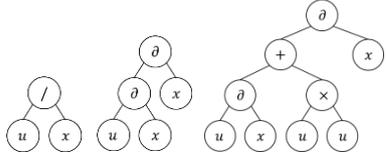 | {/ u x} & <br> {∂ (∂ u x) x} & <br> {∂ [+ (∂ u x) (× u u)] x} | $\frac{u}{x}$ & $\frac{\partial}{\partial x}\left(\frac{\partial u}{\partial x}\right)$ & $\frac{\partial}{\partial x}\left(\frac{\partial u}{\partial x} + u \cdot u\right)$ |

**2.2 Random generation of open-form PDEs**

In order to find the governing equation that best matches the data, SGA-PDE not only needs to represent any function term as a binary tree, but also needs to be able to randomly generate PDEs that comply with mathematical rules. In other words, the generated forest needs to meet the following five rules: (1) the leaf nodes of each tree are operands, and internal nodes must be operators; (2) the number of child nodes of internal nodes must be the same as the degree (i.e., the degree of all internal nodes is full); (3) the root node of each tree does not contain addition and subtraction; (4) the depth of any branch shall not exceed the given maximum depth (to avoid generating excessively complex nesting forms); and (5) the number of randomly generated trees in the forest is less than the given maximum width (to avoid generating too many redundant terms). The first three rules ensure that the generated tree is mathematically reasonable, while the latter two rules embody Occam's Razor (i.e., the simpler the better), and the purpose is to produce a concise equation structure.

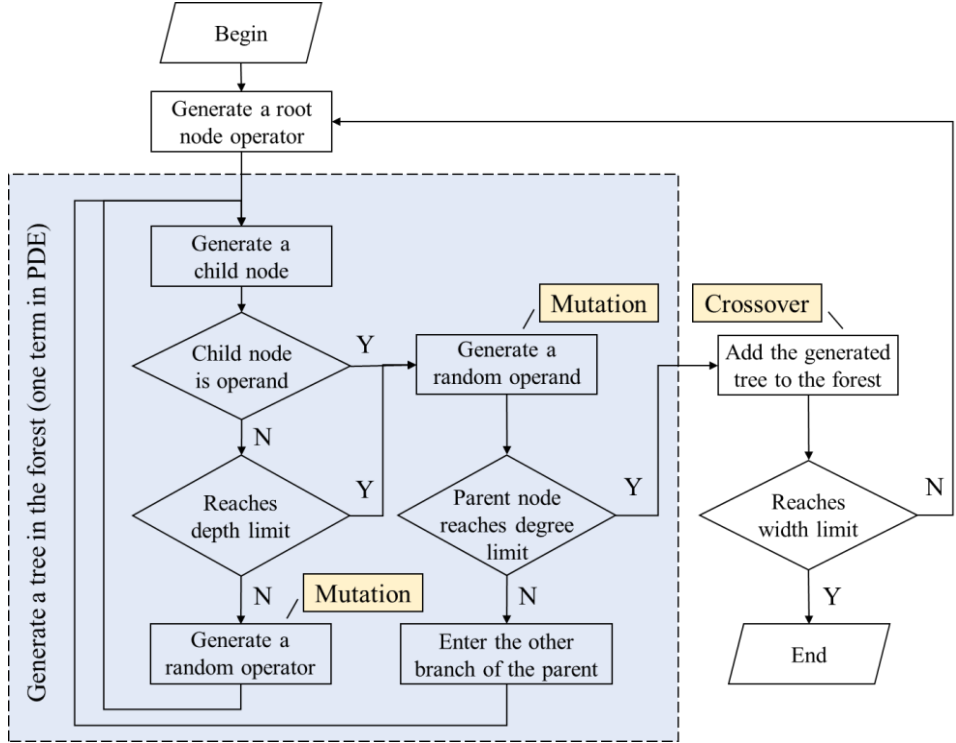

**Figure 4.** Flow chart of randomly generating a forest (PDE).

Fig. 4 shows the process of randomly generating a forest that meets the rules in SGA-PDE. First, a root node is generated for each tree in the forest, and then the child nodes of each branch of each tree are generated in turn, until a leaf node is randomly generated or the maximum depth is satisfied. Whenever the degree of an internal node is full, it enters the other branch of its parent node until the degree of the root node is also satisfied. When the degrees of all nodes of a tree are full, the current tree is complete and the next tree is ready to be generated. When the maximum number of trees (i.e., the width of PDE) is met, the random generation process ends, and a forest (PDE) with multiple trees (function terms) is obtained.

It should be noted that in the process of random generation of open-form PDEs, the mutations in the genetic algorithm introduced in the following section occur in the steps of generating operators and operands, and the crossover in the genetic algorithm happens in the process of adding new trees into the forest. The details of mutations and crossover are introduced in section 2.3.

**2.3 Genetic algorithm for binary trees**

In SGA-PDE, we use symbolic mathematics to transform any PDE into a forest of binary trees. Since all of the binary trees randomly produced by the SGA-PDE meet the mathematical rules of PDE, and any function term in a PDE can be represented as a binary tree, SGA-PDE constructs a bijection between the function term space and the binary tree space (i.e., each element of one set is paired with exactly one element of the other set, and there are no unpaired elements). This means that the open-form PDE representation method based on symbolic mathematics is effective and non-redundant, and can be used as the genetic representation of the solution domain. On this basis, this paper proposes a genetic algorithm specially designed for the tree structure to iteratively optimize candidate equations, so as to automatically mine the PDE that conforms to the observations.

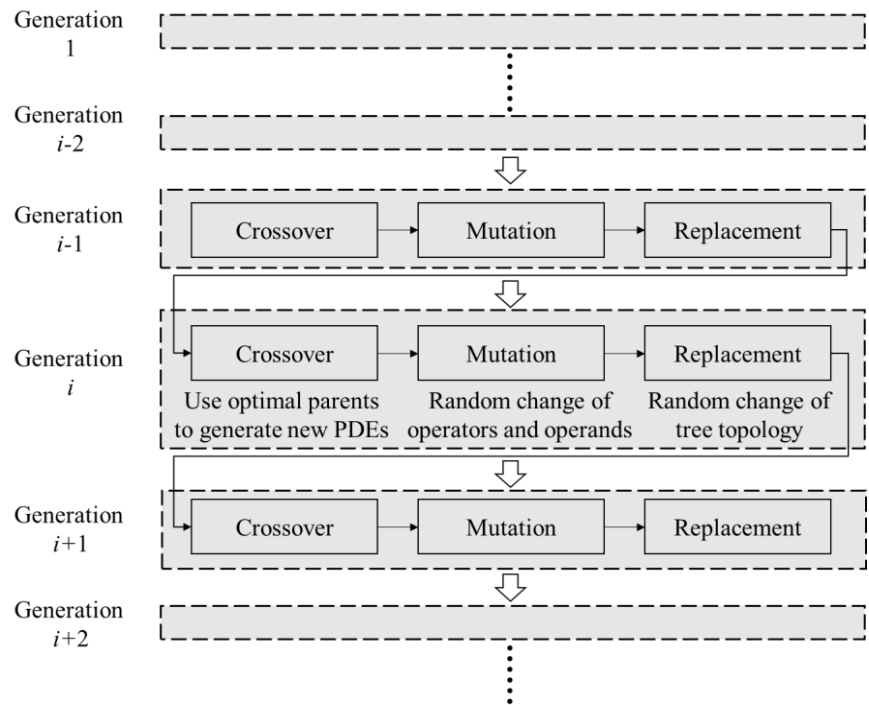

**Figure 5.** Flow chart of the genetic algorithm for the tree structure in SGA-PDE.

Specifically, SGA-PDE not only includes the crossover and mutation in the conventional genetic algorithm, but also introduces the replacement operation. SGA-PDE will first measure the fitness of each candidate solution in the current generation through the Akaike information criterion (AIC), which is taken as the performance of the current solution, and then select the best performing candidate solutions to perform crossover, mutation, and replacement operations in turn. The overall process is shown in Fig. 5.

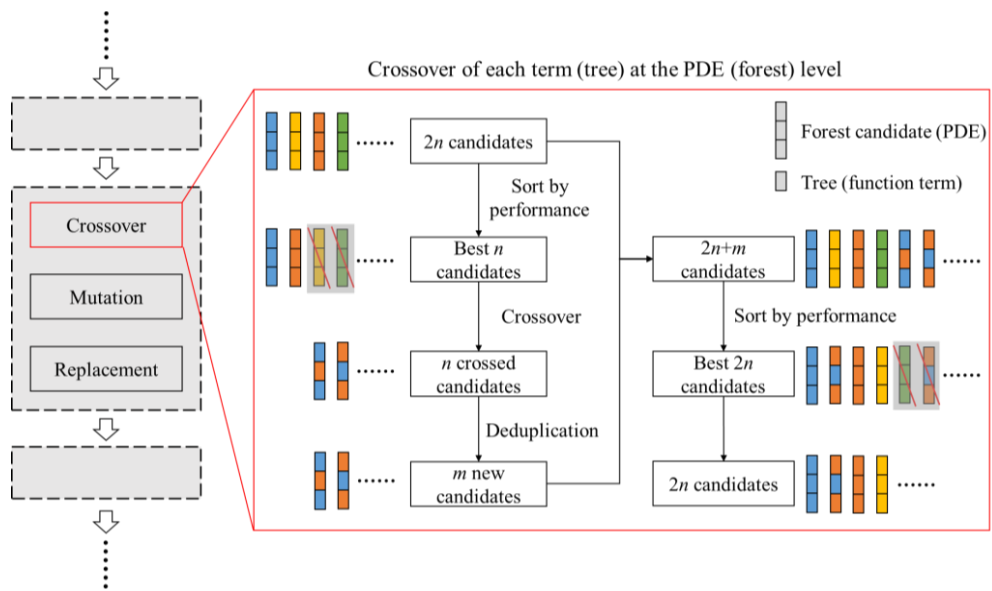

**Figure 6.** Schematic diagram of crossover in SGA-PDE.

In the crossover, the trees (function terms) in the forest (PDE) will be recombined with each other to stochastically generate new candidates from the existing population, as shown in Fig. 6. Each column in the figure represents a forest candidate (PDE), and each rectangle in the column represents a tree (function term). Assuming that there are $2n$ forest candidates in the previous generation, in the crossover of the current generation, the forest candidates are first evaluated and sorted based on AIC, and the best $n$ candidates among them are recombined to generate $n$ crossover candidates. Then, the new generated candidates are deduplicated (i.e., the candidates that have appeared in previous generations are deleted), and $m$ new candidates are obtained. Finally, according to the fitness of the data, the $2n$ candidates from the previous generation and the $m$ new candidates generated by the crossover are sorted together, and the $2n$ candidates with the best performance are taken as the result of the current generation and passed to the next step.

In the mutation, new function terms can be generated through the change of operands and operators while the topology of the existing binary tree remains unchanged. In order to ensure that the new binary tree also complies with mathematical rules of PDE, each node can only be mutated into a node with the same attribute during mutation. In other words, operands, single operators, and double operators can only be mutated into new operands, single operator, and double operator, respectively (i.e., the degree of nodes before and after the mutation is unchanged). Fig. 7 shows the mutation process of operands and operators. It can be seen that through node mutation, new function terms can be generated under the condition that the binary tree topology is completely unchanged. This also reflects the flexibility of using a binary tree to represent PDE in SGA-PDE.

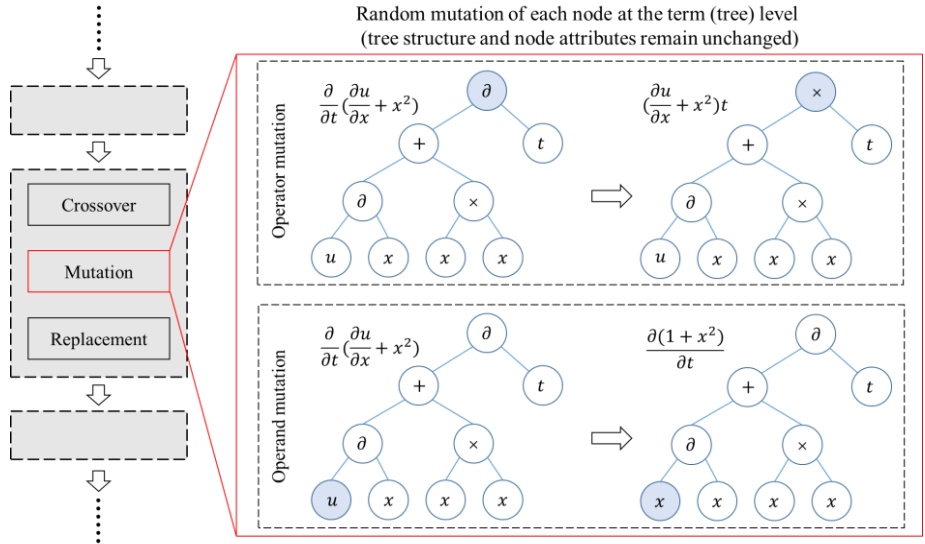

**Figure 7.** Schematic diagram of mutation in SGA-PDE.

In the replacement, a new function term is obtained by directly regenerating a tree. The mathematical meaning of a binary tree is determined by the node and the tree topological structure. The mutation generates new function terms by modifying nodes, while replacement changes the node and tree topological structure at the same time. Replacement is a more radical exploration of the solution domain and can expand the topological structure of the binary tree in the candidate set.

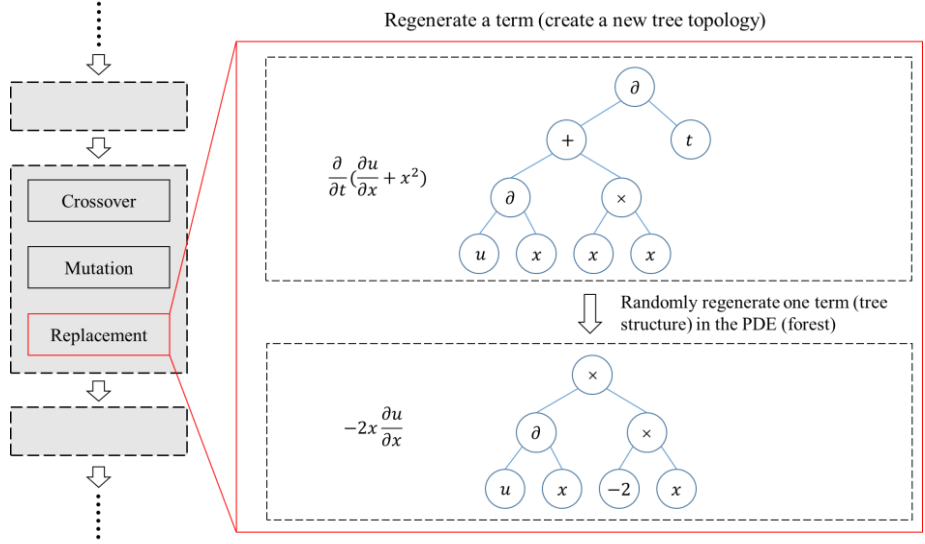

**Figure 8.** Schematic diagram of replacement in SGA-PDE.

SGA-PDE uses the topological structure of the binary tree to represent the complex function term structure in PDEs, and thus the optimization of the topological structure is an important task of the genetic algorithm. Since the mutation in SGA-PDE is to find a better node combination for a given topological structure, it is an exploration of the topological structure from the perspective of optimization. The replacement introduces new topological structures, which is essentially an exploitation of the topological structure. Therefore, through the combination of mutation and replacement, SGA-PDE can take into account the exploration and exploitation of the binary tree topology, which is conducive to efficiently finding the optimal symbolic mathematical representation of the PDE at the function term (binary tree) level.

The genetic algorithm adopted by SGA-PDE has been specially designed according to the tree structure and characteristics of PDE. Crossover is the reorganization of different binary trees in the forest, which can generate new solution candidates at the PDE level without changing the function terms (binary trees); whereas, mutation and replacement are internal changes to a certain function term, introducing new solution candidates at the function term (binary tree) level. The combination of crossover, mutation, and replacement realizes genetic variation at different levels in SGA-PDE and balances exploration and exploitation, which is important for finding the most suitable governing equation from the data.

**2.4 Determining the coefficients of PDE**

In the last step, a set of updated binary trees is obtained through the genetic algorithm (i.e., $\Phi_i(u, x)$ in Fig. 1). In this section, we use STRidge to select the binary trees that fit the observations, and construct a PDE through weighted summation of these trees. These function terms have non-zero coefficients, while other terms are removed (with zero coefficient). Finally, we evaluate the performance of the obtained PDE, and use this as a guide for iterative optimization.

STRidge is a kind of sparse regression method commonly used in PDE discovery. It was proposed by Rudy et al. in 2017[4]. PDE-FIND, DL-PDE, and DLGA-PDE are all based on this method to determine the coefficients of function terms. Specifically, the coefficients $\xi$ can be

achieved by Eq. (3):

$$\xi = \arg\min_\xi |\Phi(u,x)\cdot\xi - u_t|_2^2 + \lambda |\xi|_2^2 \tag{3}$$

In order to obtain sparse results, an appropriate threshold *tol* is introduced to select coefficients. The coefficients that are larger than *tol* are retained, while the coefficients that are smaller than *tol* are omitted. This process will continue with the remaining terms until the number of terms no longer changes. Additional details of STRidge can be found in Rudy et al.[4].

In order to ensure that the governing equation discovered by the algorithm not only fits the data but also has a concise form, the Akaike information criterion (AIC) (Eq. 4) is utilized in the SGA-PDE as a measurement of the performance of the equation:

$$AIC = 2\times k + 2\times \ln(MSE)$$
$$\text{where } MSE = \frac{|\Phi(u,x)\cdot\xi - u_t|_2^2}{N} \tag{4}$$

where $k$ denotes the number of function terms; $N$ denotes the number of observations; MSE denotes the fitness of the PDE to the data (i.e., the mean squared error between the left side and the right side of Eq. 1); $\Phi(u,x)\cdot\xi$ is calculated according to the discovered PDE; and $u_t$ is calculated based on the observations, where the derivatives are obtained by the finite difference method.

It should be noted that the PDE corresponding to the smallest AIC is not necessarily the best in PDE discovery. This is because, when the MSE is already small (i.e., the equation fits the data), introducing one or two additional compensation terms into the equation can further reduce the MSE significantly. However, this drop of MSE is not physically meaningful, i.e., it is only an overfitting of the unavoidable errors caused by various reasons, such as difference calculations. In the case of overfitting, since the penalty for the increase in the number of equation terms in AIC is not enough to offset the rapid decline of MSE, it is easy to find the equation structure with redundant terms purely based on the algorithm of simply minimizing AIC. The threshold of AIC for determining the convergence of the SGA-PDE is a hyperparameter, which may be adjusted for specific problems.

## 3. Experiment

To test the performance of the SGA-PDE, several computational experiments are carried out in this section. The main purposes of the experiments are: (1) to analyze the capability of the SGA-PDE to discover accurate governing equations from observations without any prior knowledge on the possible expression of the PDE; (2) to verify that SGA-PDE can generate a solution space covering complex equations (e.g., compound function or fractional structure) based on a limited candidate set through symbolic mathematics; and (3) to comprehensively understand the SGA-PDE, and especially demonstrate the evolution process of different PDEs (forests).

### 3.1 Problem setting

In the following subsections, we discover five different PDEs from data via SGA-PDE. The first three PDEs are governing equations with clear physical backgrounds (Burgers' equation, KdV equation, Chafee-Infante equation), and the latter two are equations with complex structures (open-form PDE with compound function or fractional structure) that cannot be handled by conventional methods. The details about the PDEs are introduced in section 3.1.1 to section 3.1.5.

**3.1.1 Burgers' equation**

Burgers' equation is a fundamental partial differential equation derived by Bateman in 1915[19, 20] and later studied by Burgers in 1948[21]. It can simulate the propagation and reflection of shock waves, which is common in a variety of dynamical problems, such as nonlinear acoustics[22] and gas dynamics[23]. The general form of Burgers' equation takes the following form:

$$\frac{\partial u}{\partial t} = -u\frac{\partial u}{\partial x} + v\frac{\partial^2 u}{\partial x^2} \tag{5}$$

where $u(x,t)$ is a given field with spatial ($x$) and temporal ($t$) dimension; and $v$ represents the diffusion coefficient or kinematic viscosity according to the problem, and it is a constant physical property.

Burgers' equation involves nonlinear terms, and it also includes an interaction term of $u$ and $\partial u/\partial x$. Therefore, this equation is used to validate the approach's effectiveness of discovering nonlinear terms from observations[4, 11]. The conventional spectral method is utilized to generate the dataset, where the diffusion coefficient $v$ is set as one[11, 24]. In the dataset, there are 201 temporal observation steps at intervals of 0.05, and 256 spatial observation steps at intervals of 0.0625. Therefore, the total number of data points is 51,456.

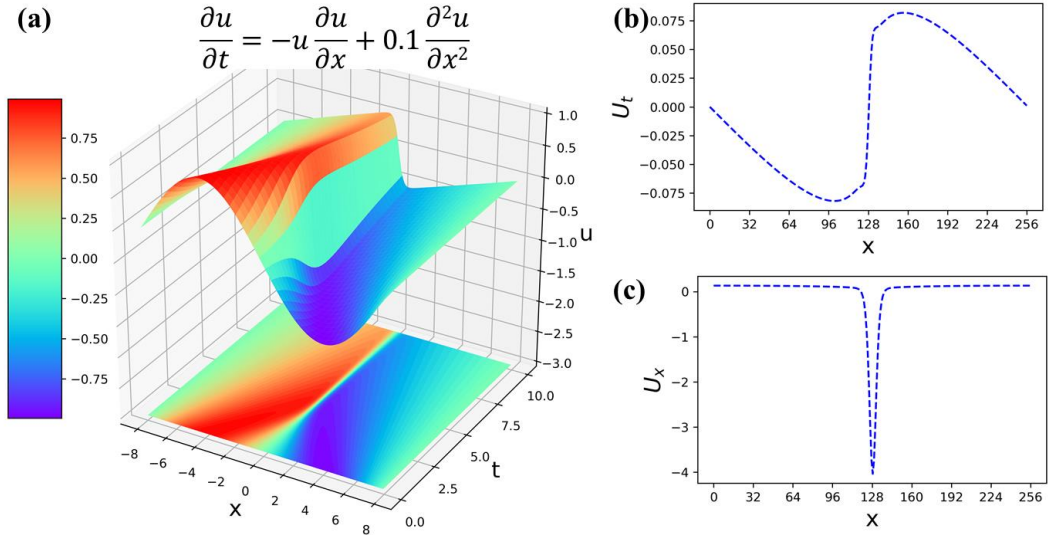

**Figure 9.** Data distribution of the observations used to mine the Burgers' equation. a: observation $u$; b: derivative $u_t$ of the observation $u$ with respect to time at all positions at the intermediate moment; c: derivative $u_x$ of the observation $u$ with respect to space at all positions at the intermediate moment.

Fig. 9a shows the distribution of the observations $u$ in Burgers' equation. The bottom of the figure is the projection of $u$ on the $x$-$t$ plane. It can be seen that Burgers' equation is volatile in space due to the influence of derivatives. In addition, the amplitude of observations will decay as time progresses. Fig. 9b and Fig. 9c, respectively, show the first derivative of the observations with respect to time ($u_t$) and space ($u_x$) at the intermediate time step ($t = 5$). It should be noted that the three variables in Fig. 9 are all of the data used by SGA-PDE when mining Burgers' equation. The data do not cover all of the function terms in Burgers' equation. Therefore, this experiment can verify the ability of SGA-PDE to generate new function terms through genetic algorithms designed for the tree structure.

### 3.1.2 Korteweg-de Vries (KdV) equation

The Korteweg-de Vries (KdV) equation is another classical example to verify the performance of the PDE discovery algorithm[4, 11]. It models the propagation of waves on shallow water surfaces. It is known as not only a prototypical example of exactly solvable models, but also one of the earliest models that have soliton solutions. The KdV equation was first discovered by Boussinesq in 1877[25], and later developed by Korteweg and de Vries in 1895 when investigating small-amplitude and long-wave motion in shallow water[26]. The KdV equation takes the following form:

$$\frac{\partial u}{\partial t} = au\frac{\partial u}{\partial x} + b\frac{\partial^3 u}{\partial x^3} \qquad (6)$$

where $a$ and $b$ are constants, which are set as $a = -1$ and $b = -0.0025$, respectively.

Since the KdV equation contains a third-order derivative term, its volatility is greater and it is challenging to learn the correct equation structure directly from the data. In order to increase the difficulty of the experiment, we only provide the original observation $u$ (Fig. 10a) and the first derivative of $u$ with respect to time and space (Fig. 10b and Fig. 10c) for the SGA-PDE. The ability of SGA-PDE to find higher-order derivatives that do not exist in the candidate set based on limited data can be verified through this experiment. The conventional spectral method is utilized to generate the dataset[11]. Regarding the data size, there are 201 temporal observation steps at intervals of 0.005, and 512 spatial observation steps at intervals of 0.0039. Therefore, the total number of data points is 102,912.

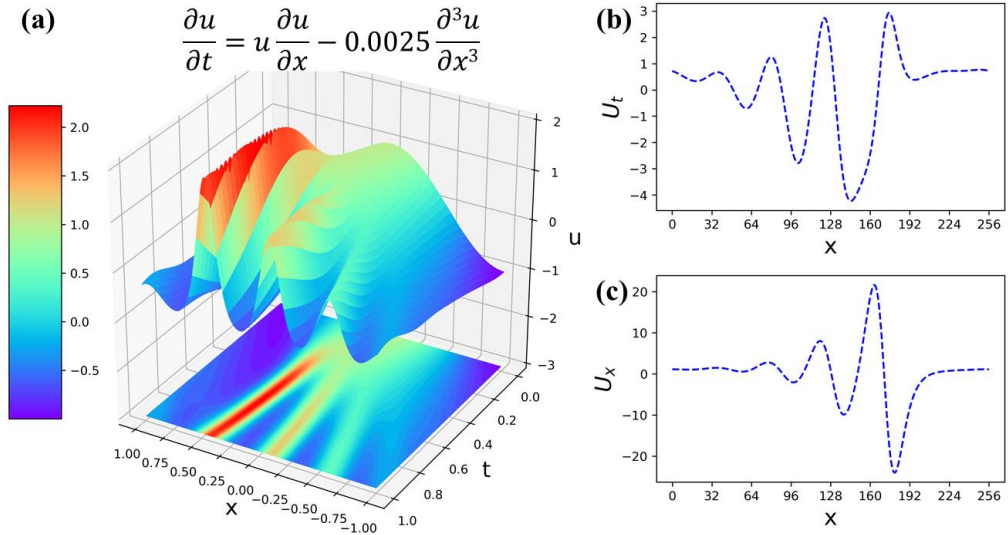

**Figure 10.** Data distribution of the observations used to mine the KdV equation. a: observation $u$; b: $u_t$ at all positions at the intermediate moment; c: $u_x$ at all positions at the intermediate moment.

### 3.1.3 Chafee-Infante equation

The Chafee-Infante equation is a reaction diffusion equation proposed by Chafee and Infante in 1974[27], which describes the physical processes of material transport and particle diffusion, and it is applied in fluid mechanics, high-energy physical processes, electronics, and environmental science[28, 29]. It is a kind of nonlinear evolution equation in the form of Eq. 7:

$$\frac{\partial u}{\partial t} = \frac{\partial^2 u}{\partial x^2} + a\left(u - u^3\right) \tag{7}$$

where $a$ is the diffusion coefficient, which is set as $a = 1$ in this study.

The Chafee-Infante equation is also called the Newell-Whitehead equation when the diffusion coefficient $a$ is 1[30, 31]. The Chafee-Infante equation has a second-order derivative and a third-order exponential term, with strong nonlinearity. The dataset is generated via the forward difference method[11], and there are 200 temporal observation steps at intervals of 0.002, and 301 spatial observation steps at intervals of 0.01. Therefore, the total number of data points is 60,200. All of the data used by SGA-PDE to discover the Chafee-Infante equation is shown in Fig. 11.

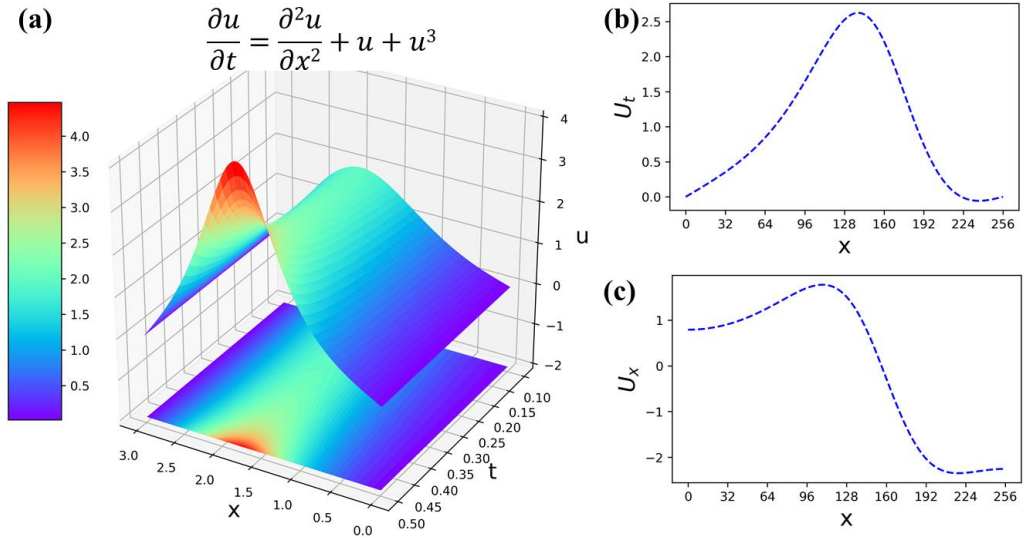

**Figure 11.** Data distribution of the observations used to mine the Chafee-Infante equation. a: observation $u$; b: $u_t$ at all positions at the intermediate moment; c: $u_x$ at all positions at the intermediate moment.

### 3.1.4 Open-form PDE with fractional structure and compound function

In order to verify the ability of SGA-PDE to mine complex open-form PDEs, we generated two testing datasets based on two PDEs. The first PDE has a fractional structure (PDE_divide), as shown in Eq. 8. The second is PDE_compound, which contains a compound function (Eq. 9).

$$\frac{\partial u}{\partial t} = \frac{-1}{x}\frac{\partial u}{\partial x} + 0.25\frac{\partial^2 u}{\partial x^2} \tag{8}$$

$$\frac{\partial u}{\partial t} = \frac{\partial}{\partial x}\left(u\frac{\partial u}{\partial x}\right) \tag{9}$$

Specifically, to generate datasets of PDE_divide and PDE_compound, the problems are solved numerically using the finite difference method, where $x \in [1,2]$. The initial condition is $u(x,0) = -\sin(\pi x)$, and the boundary condition is $u(1,t) = u(2,t) = 0, t > 0$. Both PDEs are solved with 100 spatial observation points for 250,001 timesteps. Then, temporal observation points are taken every 1,000 timesteps. Therefore, we have 100 spatial observation points and 251 temporal observation points, and the total number of the dataset is 25,100.

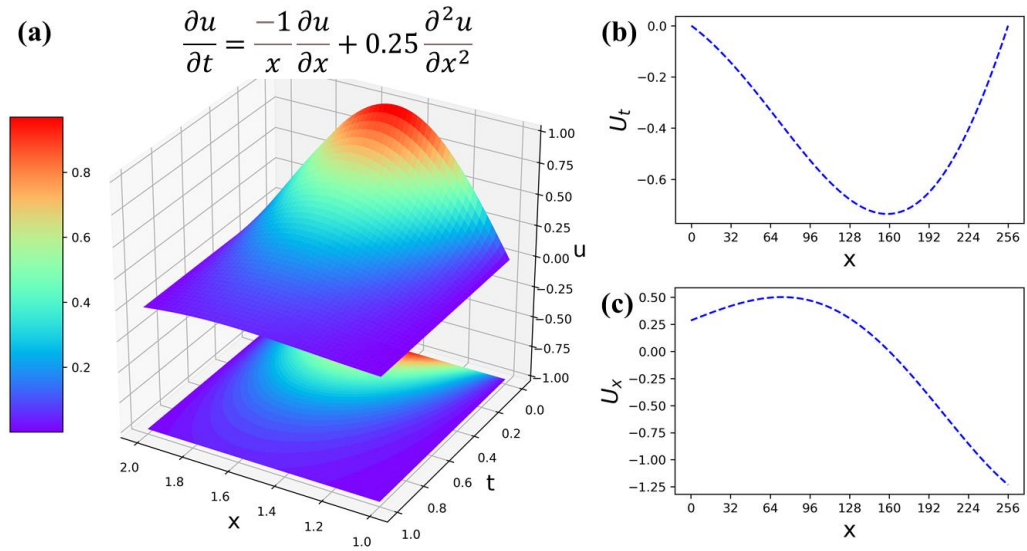

**Figure 12.** Data distribution of the observations used to mine the PDE with fractional structure. a: observation $u$; b: $u_t$ at all positions at the intermediate moment; c: $u_x$ at all positions at the intermediate moment.

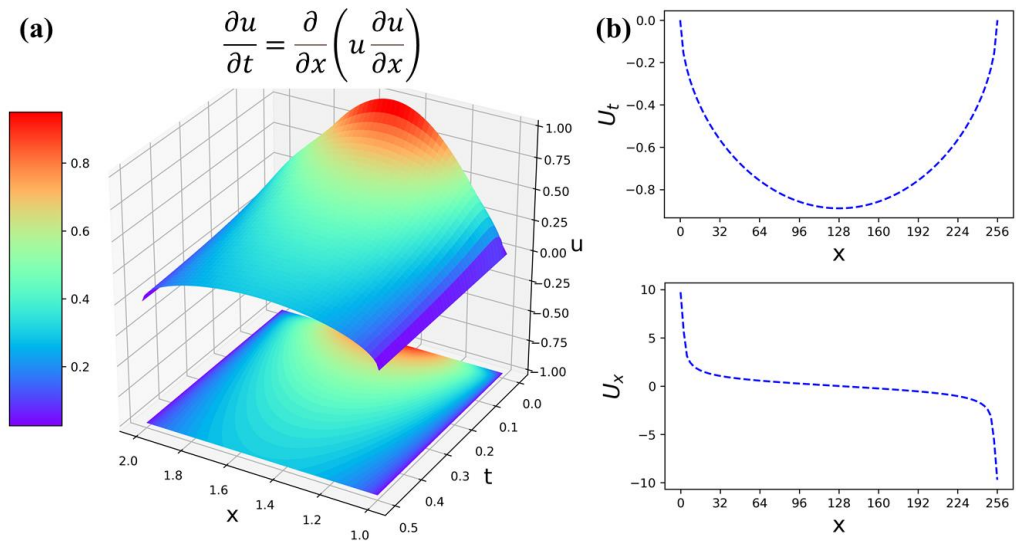

**Figure 13.** Data distribution of the observations used to mine the PDE with compound function. a: observation $u$;

b: $u_t$ at all positions at the intermediate moment; c: $u_x$ at all positions at the intermediate moment.

Since it is almost impossible to determine all potential function terms with fractional structure or compound functions in the candidate set in advance, PDE discovery algorithms based on sparse regression (e.g., PDE-FIND and DL-PDE) cannot handle the open-form PDEs, such as Eq. 8 and Eq. 9. Although DLGA based on the genetic algorithm can extend the candidate set, it still lacks a method to generate compound functions or fractional structures through a simple combination of gene fragments. Therefore, the above two open-form PDEs are challenging for conventional methods.

Finally, the generated dataset of PDE_divide is shown in Fig. 12, and the data distribution of PDE_compound is shown in Fig. 13. Although the changing trends of the observation values of the two PDEs appear similar, their governing equations are completely dissimilar, which also reflects the difficulty of directly mining the control equations through data. These two equations can verify the SGA-PDE's ability to mine open-form PDEs through its flexible representation, which is essential for the future use of SGA-PDE to extract unknown and undiscovered governing equations from data.

**3.2 Performance of SGA-PDE**

In the following five experiments, SGA-PDE used the same operators and operands. Specifically, the double operators include addition $+$, subtraction $-$, multiplication $\times$, division $\div$, first-order differential $\partial$, and second-order differential $\partial^2$; the single operators are square $()^2$ and cube $()^3$; the operands are the dependent variable $u$, independent variable $x$, first derivative $u_x$, and zero 0. The zero in the operand plays the role of pruning in the tree by muting a certain leaf/branch. The zero operand is an important device in SGA-PDE, which is helpful for finding the optimal node combination when the topology of the binary tree remains unchanged. Moreover, adding the dependent variable $u$ into the candidate set is a strategy to speed up the optimization in SGA-PDE. This is because the dependent variable is a common function term in PDE, and thus it is added as a default candidate tree in the experiments. Since the operator contains division, in order to avoid generating mathematically meaningless solutions, when the denominator (i.e., the right child node of division) is equal to zero, it is replaced by $10^{-10}$. Furthermore, the left side of the PDE ($u_t$) is also provided in this study, and thus the core task is to find the correct form on the right side of the PDE through SGA-PDE. This restriction may be relaxed easily[11]. In order to verify the generalization of SGA-PDE, the operators and operands are not adjusted according to the problem in the experiments. In addition, since the operators are basic, there is no need to make many assumptions about the equation structure in SGA-PDE, which indicates the ability of SGA-PDE to discover new equations from the data without prior knowledge of the mechanism of the problem.

The same hyperparameters are used in all five PDE experiments. Specifically, the maximum generation is set to 100, and the population of each generation is 20. The probability of generating a node as an operand (leaf node) instead of an operator is 0.5. The probability of mutation at each node is 0.3. The probability of crossover between different function terms in two PDE is 0.5. Moreover, SGA-PDE also contains the maximum forest width hyperparameter (which is set to five) to constrain the number of function terms in the PDE, and the maximum tree depth hyperparameter (which is set to four) to constrain the number of nesting calculations in each function term. The larger is the maximum tree depth, the more likely it is to produce nested function terms. The AIC threshold is set to -10.

Finally, SGA-PDE successfully discovers the correct governing equations of all five problems, as shown in Table 3. The experimental results show that SGA-PDE can not only discover the PDEs that can be found by sparse regression methods (e.g., Burgers' equation of nonlinear interaction terms and KdV equation with high-order derivatives), and the PDEs that can be solved by genetic algorithms (e.g., Chafee-Infante equation with both high-order derivatives and high-order exponents), but also handle the complex open-form PDEs (e.g., PDE_divide with fractional structure and PDE_compound with compound function). The latter two PDEs are difficult to be mined directly from data by conventional methods. Nevertheless, the SGA-PDE successfully handles them due to the flexibility brought by its flexible symbolic representation method.

Table 3. Experimental results of different PDE discovery models mining governing equations from data.

| | Correct equation structure | PDE-FIND[4] | PDE-net[9] | DL-PDE[6] | EPDE[10] | DLGA-PDE[11] | **SGA-PDE** |
|---|---|---|---|---|---|---|---|
| Burgers' | $\frac{\partial u}{\partial t} = -u\frac{\partial u}{\partial x} + 0.1\frac{\partial^2 u}{\partial x^2}$ | ✓ | ✓ | ✓ | ✓ | ✓ | ✓ |
| KdV | $\frac{\partial u}{\partial t} = -0.0025\frac{\partial^3 u}{\partial x^3} - u\frac{\partial u}{\partial x}$ | ✓ | | ✓ | ✓ | ✓ | ✓ |
| Chafee-Infante | $\frac{\partial u}{\partial t} = \frac{\partial^2 u}{\partial x^2} - u + u^3$ | | | | | ✓ | ✓ |
| PDE_divide | $\frac{\partial u}{\partial t} = \frac{-1}{x}\frac{\partial u}{\partial x} + 0.25\frac{\partial^2 u}{\partial x^2}$ | | | | | | ✓ |
| PDE_compound | $\frac{\partial u}{\partial t} = \frac{\partial}{\partial x}(u\frac{\partial u}{\partial x})$ | | | | | | ✓ |

In order to comprehensively understand the SGA-PDE, Fig. 14 shows the best binary trees and the corresponding equation in each generation to demonstrate the evolution process of finding the KdV equation. It should be noted that when the value of the right child node of the division operator is equal to zero, it will be replaced with $10^{-10}$ to avoid mathematically meaningless solutions. It can be seen from Fig. 14 that among the 20 randomly generated samples of the first generation, the equation structure that best fits the observations is $(u - x)(u - \frac{\partial u}{\partial x})$, which is obviously far from the real solution. In the subsequent evolution process, with the mutation of the nodes and topological structure of the binary trees, as well as the crossover between the function terms, the correct solution is finally found in the 31st generation: $\frac{\partial^3 u}{\partial x^3}$ and $u\frac{\partial u}{\partial x}$, where the corresponding AIC is -10.3 and meets the convergence threshold. If the genetic optimization continues, although the tree structure becomes more complicated, the solutions do not change in essence (i.e., the equivalent form of the KdV equation). In addition, there is an interesting phenomenon in the 69th generation: the SGA-PDE discovers a more concise expression of $u\frac{\partial u}{\partial x}$ based on the derivation of a compound function, i.e., $0.5\frac{\partial u^2}{\partial x}$.

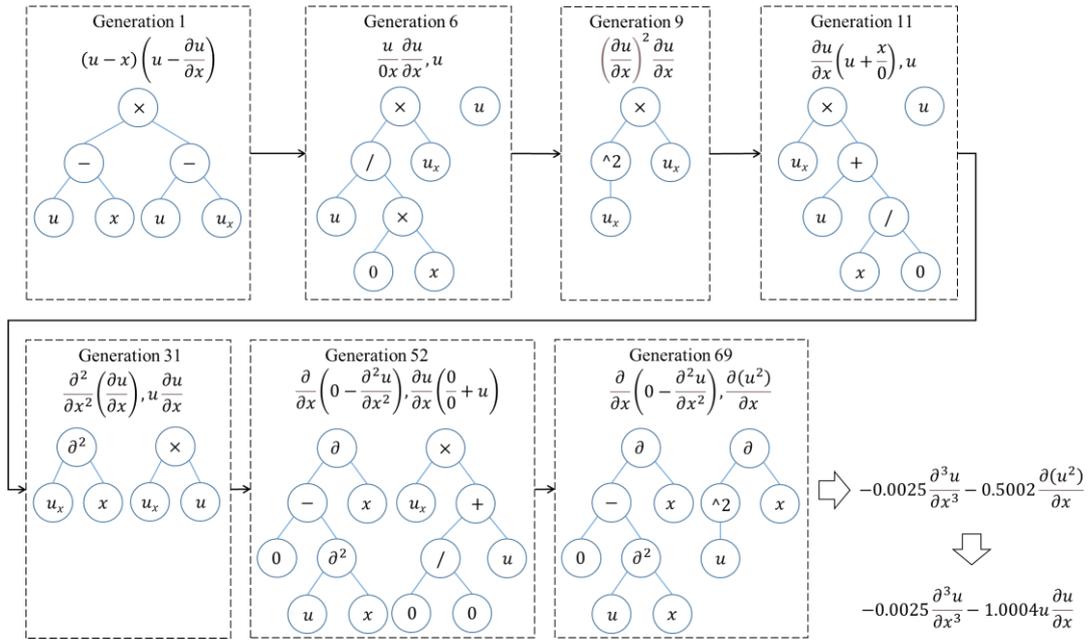

**Figure 14.** Diagram of best binary trees in each generation in the evolution process of SGA-PDE mining KdV equation.

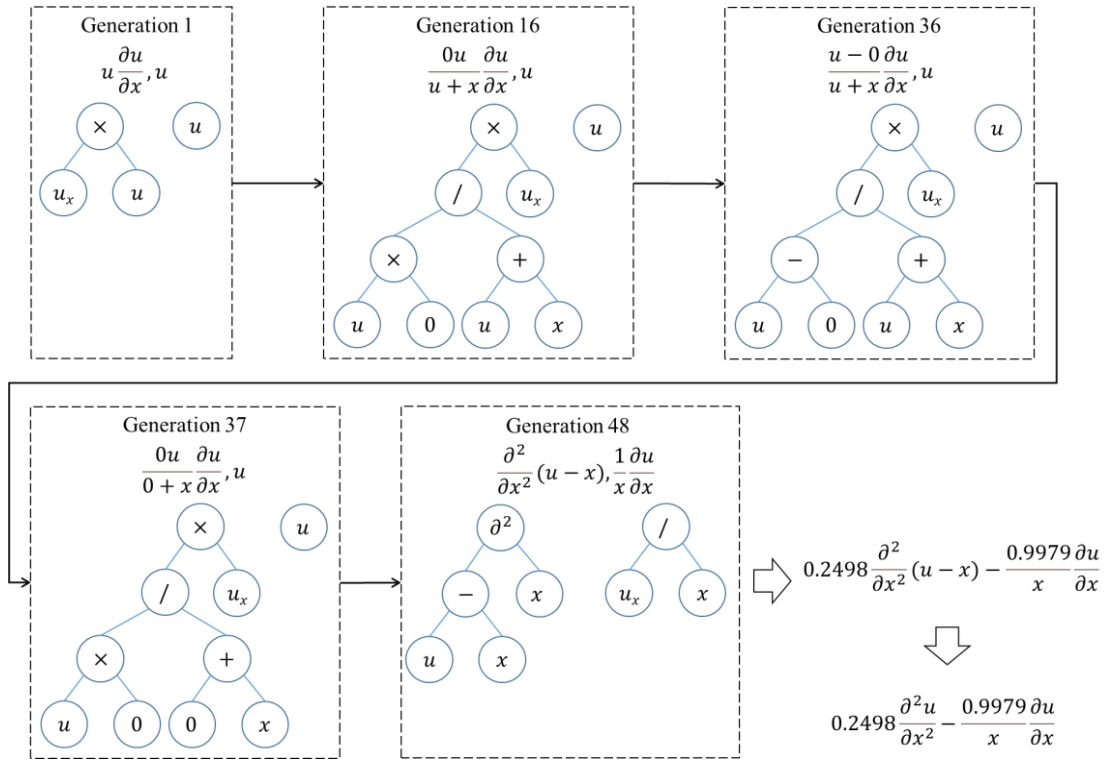

**Figure 15.** Diagram of best binary trees in each generation in the evolution process of SGA-PDE mining PDE_divide.

Fig. 15 and Fig. 16 show the evolution process of SGA-PDE searching for PDEs with fractional structures and compound functions, which cannot be handled by conventional methods. The figures only show the generation of the optimal binary tree that has changed. However, mutation and optimization of the non-optimal binary trees are still ongoing in the generations that are not shown in the figures, which is the basis for new optimal binary tree in future generations. Due to the flexible representation method of the binary tree, SGA-PDE can effectively find function terms with fractional structures (e.g., $\frac{1}{x}\frac{\partial u}{\partial x}$) and function terms with compound functions (e.g., $\frac{\partial}{\partial x}(u\frac{\partial u}{\partial x})$). Furthermore, in the process of mining PDE_compound from the data, SGA-PDE found the concise equivalent form ($0.5\frac{\partial^2(u^2)}{\partial x^2}$) of the correct function term.

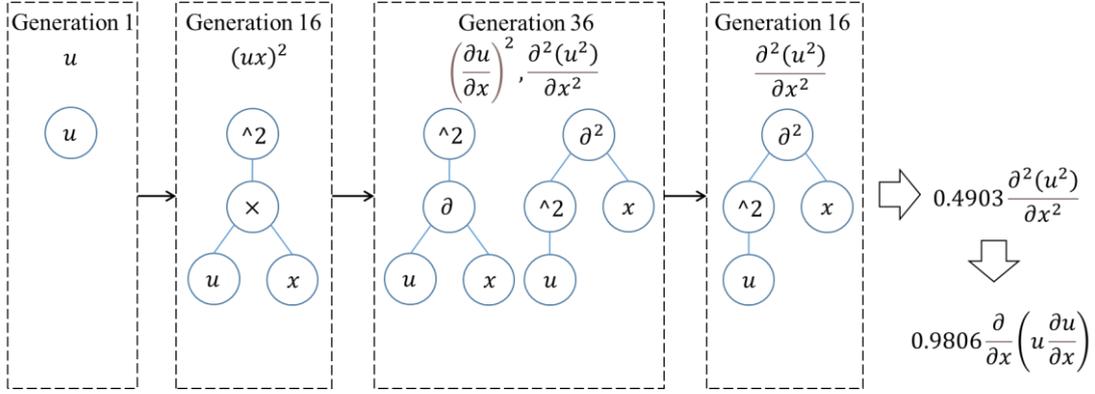

Figure 16. Diagram of best binary trees in each generation in the evolution process of SGA-PDE mining PDE_compound.

In order to quantitatively evaluate the performance of SGA-PDE, Table 4 compares the reference PDE and the PDE found by the SGA-PDE, and calculates the fitness between the discovered PDE and the observations. It can be seen that the function terms of the PDE found by SGA-PDE are completely consistent with the reference PDE, which verifies that SGA-PDE has the ability to directly mine open-form governing equations directly from data.

Table 4. Comparison of reference PDE and the PDE found by SGA-PDE.

|  | Reference PDE | SGA-PDE discovered PDE | MSE |
| --- | --- | --- | --- |
| Burgers' | $\frac{\partial u}{\partial t} = -u\frac{\partial u}{\partial x} + 0.1\frac{\partial^2 u}{\partial x^2}$ | $\frac{\partial u}{\partial t} = -1.0011u\frac{\partial u}{\partial x} + 0.1024\frac{\partial^2 u}{\partial x^2}$ | $4.33\times 10^{-5}$ |
| KdV | $\frac{\partial u}{\partial t} = -0.0025\frac{\partial^3 u}{\partial x^3} - u\frac{\partial u}{\partial x}$ | $\frac{\partial u}{\partial t} = -0.0025\frac{\partial^3 u}{\partial x^3} - 1.0004u\frac{\partial u}{\partial x}$ | $1.48\times 10^{-4}$ |
| Chafee-Infante | $\frac{\partial u}{\partial t} = \frac{\partial^2 u}{\partial x^2} - u + u^3$ | $\frac{\partial u}{\partial t} = 1.0002\frac{\partial^2 u}{\partial x^2} - 1.0008u + 1.0004u^3$ | $4.84\times 10^{-5}$ |
| PDE_divide | $\frac{\partial u}{\partial t} = \frac{-1}{x}\frac{\partial u}{\partial x} + 0.25\frac{\partial^2 u}{\partial x^2}$ | $\frac{\partial u}{\partial t} = \frac{-0.9979}{x}\frac{\partial u}{\partial x} + 0.2498\frac{\partial^2 u}{\partial x^2}$ | $1.78\times 10^{-4}$ |
| PDE_compound | $\frac{\partial u}{\partial t} = \frac{\partial}{\partial x}(u\frac{\partial u}{\partial x})$ | $\frac{\partial u}{\partial t} = 0.9806\frac{\partial}{\partial x}(u\frac{\partial u}{\partial x})$ | $1.13\times 10^{-1}$ |

## 5. Conclusion and discussion

This study proposes the symbolic genetic algorithm (SGA-PDE) to automatically mine the PDE (governing equation) from data by processing equations as forests of binary trees. The key to finding the PDE that best fits the data lies in solving two problems: (1) representation: how to represent any given complex open-form PDEs through symbolic mathematics; and (2) optimization: how to use machine learning algorithms to mine the correct equation from the infinite feasible region of PDEs.

Firstly, SGA-PDE uses binary trees to build a flexible representation of PDEs with complex structures. By introducing the concept of operands and operators, SGA-PDE transforms the representation units of the equations from the function term level to the operator and operand level, which makes SGA-PDE more flexible and capable to handle complex open-form PDEs compared with conventional methods. Ultimately, any partial differential governing equation can be represented as a forest, and the function terms in the PDE are transformed as binary trees in SGA-PDE. It should be mentioned that although the flexible binary trees enable SGA-PDE to represent any equation, the gradient between the binary trees and the loss (i.e., the misfit between PDE and the observations) is challenging to calculate, which makes it difficult to use conventional gradient-based optimization methods to find the PDE that best fits the observations.

Secondly, in order to take advantage of the flexible symbolic representation, a gradient-free optimization method is required. This study proposes a specially designed genetic algorithm for binary trees to iteratively optimize the forests (PDEs), and realizes the efficient optimization of the open-form PDE without prior knowledge of the possible expression of the equations. The proposed genetic algorithm for binary trees includes three parts: crossover (forest level optimization), mutation, and replacement (binary tree level optimization). The crossover generates new solutions by exchanging function terms with different PDEs. The mutation is to find a better node combination for a given tree topological structure to fit the observations, and thus it is an exploration of the topological structure from the perspective of optimization. Replacement produces new topological structures, which is essentially an exploitation. Overall, the genetic algorithm for binary trees can not only simultaneously optimize the solutions at the PDE level and the function term level, but also balance the exploration and exploitation of the topological structure of the binary trees.

In the experiment, the performance of SGA-PDE is examined by mining five governing PDEs with strong nonlinearity directly from data, among which Burgers' equation has interaction terms, KdV equation has high-order derivatives, Chafee-Infante equation has derivative and exponential terms, PDE_divide has a fractional structure, and PDE_compound contains a compound function in the derivative term. The last two study cases cannot be handled by conventional methods. Indeed, SGA-PDE successfully solved all of the study cases. These experiments show that SGA-PDE benefits from its flexible representation of PDE and genetic algorithm for trees, and it can discover and optimize PDE without any prior knowledge of the expression of the PDE.

It should be noted that the SGA-PDE does not require a candidate set that covers all possible function terms. Therefore, SGA-PDE has the potential to explore more complex PDEs that are unknown in real problems. Although SGA-PDE does not rely on prior information, when the prior information or domain knowledge is available, SGA-PDE can also utilize them through the preset candidate set. Furthermore, in recent years, many researchers have used scientific machine learning to embed governing PDEs into neural networks (i.e., knowledge embedding)[32], and have improved model performance in numerous problems[24, 33-35]. However, many practical problems lack governing equations, which limits the application of knowledge embedding in practice. SGA-PDE is a kind of knowledge discovery method, which has the potential to solve the problem of a lack of governing equations in real-world scenarios. Therefore, SGA-PDE can support the development of scientific machine learning by discovering governing equations from experimental and simulation data, and transfer them to models in other scenarios. In addition, the discovered PDE is interpretable, which is conducive to deepening our understanding of the physical world.

In future studies, SGA-PDE will be used to solve unknown problems in real-world scenarios. Data in the real world are typically sparse and noisy. We will explore more methods to improve the robustness and the ability of small data learning of SGA-PDE, in order to better apply it to real-world problems. SGA-PDE has the potential to discover unknown governing equations directly from data, and the equations can be further integrated into neural networks through knowledge embedding methods to improve the performance and robustness of the model. PDE discovery and knowledge embedding can form a closed loop in scientific machine learning, which will be explored in future studies.

**Acknowledgements**

This work is partially funded by the Shenzhen Key Laboratory of Natural Gas Hydrates (Grant No. ZDSYS20200421111201738) and the SUSTech --- Qingdao New Energy Technology Research Institute.

**References**

1. Binder, K., Heermann, D., Roelofs, L., Mallinckrodt, A.J., and McKay, S., *Monte Carlo simulation in statistical physics.* Computers in Physics, 1993. **7**(2): p. 156-157.
2. Griebel, M., Dornseifer, T., and Neunhoeffer, T., *Numerical Simulation in Fluid Dynamics: A Practical Introduction*. 1998: SIAM.
3. Pletcher, R.H., Tannehill, J.C., and Anderson, D., *Computational Fluid Mechanics and Heat Transfer*. 2012: CRC press.
4. Rudy, S.H., Brunton, S.L., Proctor, J.L., and Kutz, J.N., *Data-driven discovery of partial differential equations.* Science Advances, 2017. **3**(4): p. e1602614.
5. Chang, H. and Zhang, D., *Machine learning subsurface flow equations from data.* Computational Geosciences, 2019. **23**(5): p. 895-910.
6. Xu, H., Chang, H., and Zhang, D., *Dl-pde: Deep-learning based data-driven discovery of partial differential equations from discrete and noisy data.* Communications in Computational Physics, 2019. **39**(3): p. 698-728.
7. Schaeffer, H., *Learning partial differential equations via data discovery and sparse optimization.* Proceedings of the Royal Society A: Mathematical, Physical Engineering Sciences, 2017. **473**(2197): p. 20160446.
8. Brunton, S.L., Proctor, J.L., and Kutz, J.N., *Discovering governing equations from data by sparse identification of nonlinear dynamical systems.* Proceedings of the National Academy of Sciences, 2016. **113**(15): p. 3932-3937.
9. Long, Z., Lu, Y., Ma, X., and Dong, B. *Pde-net: Learning pdes from data*. in *International Conference on Machine Learning*. 2018. PMLR.
10. Maslyaev, M., Hvatov, A., and Kalyuzhnaya, A. *Data-driven partial derivative equations discovery with evolutionary approach*. in *International Conference on Computational Science*.

2019. Springer.

11. Xu, H., Chang, H., and Zhang, D., *DLGA-PDE: Discovery of PDEs with incomplete candidate library via combination of deep learning and genetic algorithm.* Journal of Computational Physics, 2020. **418**: p. 109584.

12. Long, Z., Lu, Y., and Dong, B., *PDE-Net 2.0: Learning PDEs from data with a numeric-symbolic hybrid deep network.* Journal of Computational Physics, 2019. **399**: p. 108925.

13. Lample, G. and Charton, F., *Deep learning for symbolic mathematics.* arXiv preprint arXiv:.01412, 2019.

14. Liu, C., Zoph, B., Neumann, M., Shlens, J., Hua, W., Li, L.J., Li, F.F., Yuille, A., Huang, J., and Murphy, K. *Progressive neural architecture search*. in *Proceedings of the European conference on computer vision (ECCV)*. 2018.

15. Lu, Z., Whalen, I., Boddeti, V., Dhebar, Y., Deb, K., Goodman, E., and Banzhaf, W. *Nsga-net: neural architecture search using multi-objective genetic algorithm*. in *Proceedings of the Genetic and Evolutionary Computation Conference*. 2019.

16. Luo, R., Tian, F., Qin, T., Chen, E., and Liu, T.-Y., *Neural architecture optimization.* arXiv preprint arXiv:.07233, 2018.

17. Liu, H., Simonyan, K., and Yang, Y., *Darts: Differentiable architecture search.* arXiv preprint arXiv:.09055, 2018.

18. Zoph, B. and Le, Q.V., *Neural architecture search with reinforcement learning.* arXiv preprint arXiv:.01578, 2016.

19. Bateman, H., *Some recent researches on the motion of fluids.* Monthly Weather Review, 1915. **43**(4): p. 163-170.

20. Whitham, G.B., *Linear and nonlinear waves*. Vol. 42. 2011: John Wiley & Sons.

21. Burgers, J.M., *A mathematical model illustrating the theory of turbulence*, in *Advances in applied mechanics*. 1948, Elsevier. p. 171-199.

22. Rudenko, O. and Soluian, S., *The theoretical principles of nonlinear acoustics.* Moscow Izdatel Nauka, 1975.

23. Kutluay, S., Bahadir, A., and Özdeş, A., *Numerical solution of one-dimensional Burgers equation: explicit and exact-explicit finite difference methods.* Journal of Computational Applied Mathematics, 1999. **103**(2): p. 251-261.

24. Raissi, M., Perdikaris, P., and Karniadakis, G.E., *Physics-informed neural networks: A deep learning framework for solving forward and inverse problems involving nonlinear partial differential equations.* Journal of Computational Physics, 2019. **378**: p. 686-707.

25. Boussinesq, J., *Essai sur la théorie des eaux courantes*. 1877: Impr. nationale.

26. Korteweg, D.J. and De Vries, G., *On the change of form of long waves advancing in a rectangular canal, and on a new type of long stationary waves.* The London, Edinburgh, Dublin Philosophical Magazine Journal of Science, 1895. **39**(240): p. 422-443.

27. Chafee, N. and Infante, E.F., *A bifurcation problem for a nonlinear partial differential equation of parabolic type.* Applicable Analysis, 1974. **4**(1): p. 17-37.

28. Tahir, M., Kumar, S., Rehman, H., Ramzan, M., Hasan, A., and Osman, M.S., *Exact traveling wave solutions of Chaffee–Infante equation in (2+ 1)-dimensions and dimensionless Zakharov equation.* Mathematical Methods in the Applied Sciences, 2021. **44**(2): p. 1500-1513.

29. Sun, Y.-h., Yang, S.-h., Wang, J., and Liu, F.-s., *New exact solutions of nonlinear Chafee-Infante reaction and diffusion equation.* Journal of Sichuan Normal University, 2012. **3**.

30. Newell, A.C. and Whitehead, J.A., *Finite bandwidth, finite amplitude convection.* Journal of Fluid Mechanics, 1969. **38**(2): p. 279-303.
31. Korkmaz, A., *Complex wave solutions to mathematical biology models I: Newell–Whitehead–Segel and Zeldovich equations.* Journal of Computational Nonlinear Dynamics, 2018. **13**(8).
32. Baker, N., Alexander, F., Bremer, T., Hagberg, A., Kevrekidis, Y., Najm, H., Parashar, M., Patra, A., Sethian, J., and Wild, S., *Workshop report on basic research needs for scientific machine learning: Core technologies for artificial intelligence*. 2019, USDOE Office of Science (SC), Washington, DC (United States).
33. Raissi, M., Yazdani, A., and Karniadakis, G.E., *Hidden fluid mechanics: Learning velocity and pressure fields from flow visualizations.* Science, 2020. **367**(6481): p. 1026-1030.
34. Chen, Y., Huang, D., Zhang, D., Zeng, J., Wang, N., Zhang, H., and Yan, J., *Theory-guided hard constraint projection (HCP): a knowledge-based data-driven scientific machine learning method.* arXiv preprint arXiv:.06148, 2020.
35. Wang, N., Zhang, D., Chang, H., and Li, H., *Deep learning of subsurface flow via theory-guided neural network.* Journal of Hydrology, 2020. **584**: p. 124700.